\pdfoutput=1

\documentclass[11pt]{article}

\usepackage{emnlp2021}

\usepackage{times}
\usepackage{latexsym}

\usepackage[T1]{fontenc}

\usepackage[utf8]{inputenc}

\usepackage{microtype}

\usepackage{xcolor}

\usepackage{graphicx}
\usepackage{amsmath}
\usepackage{bm}
\usepackage{amssymb}
\usepackage{bbding}
\usepackage{multicol,multirow}
\usepackage{tabularx,booktabs}       
\newcolumntype{Y}{>{\hsize=.7\hsize\centering\arraybackslash}X}
\newcolumntype{s}{>{\hsize=.25\hsize\centering\arraybackslash}X}
\newcolumntype{m}{>{\hsize=.001\hsize}X}
\graphicspath{ {./images/} }

\title{Ultra-High Dimensional Sparse Representations with Binarization for Efficient Text Retrieval}

\author{Kyoung-Rok Jang\textsuperscript{1}, Junmo Kang\textsuperscript{1}\thanks{\ \ These authors contributed equally.} , Giwon Hong\textsuperscript{1}\footnotemark[1] , Sung-Hyon Myaeng\textsuperscript{1}\thanks{\hspace{0.15cm}The corresponding author}, \\ {\bf Joohee Park\textsuperscript{2}, Taewon Yoon\textsuperscript{2}, Heecheol Seo\textsuperscript{2}} \\
        \textsuperscript{1}KAIST, Republic of Korea \\
        \textsuperscript{2}NAVER, Republic of Korea \\
        \tt{\{kyoungrok.jang, junmo.kang, giwon.hong, myaeng\}@kaist.ac.kr} \\
        \tt{\{james.joohee.park, taewon.yoon, heecheol.seo\}@navercorp.com} 
}

\begin{document}
\setlength{\abovedisplayskip}{-5pt}
\setlength{\belowdisplayskip}{5pt}
\setlength{\abovedisplayshortskip}{-5pt}
\setlength{\belowdisplayshortskip}{2pt}








\maketitle
\begin{abstract}
The semantic matching capabilities of neural information retrieval can ameliorate synonymy and polysemy problems of symbolic approaches. However, neural models' dense representations are more suitable for re-ranking, due to their inefficiency. Sparse representations, either in symbolic or latent form, are more efficient with an inverted index. Taking the merits of the sparse and dense representations, we propose an ultra-high dimensional (UHD) representation scheme equipped with directly controllable sparsity. UHD's large capacity and minimal noise and interference among the dimensions allow for \textit{binarized} representations, which are highly efficient for storage and search. Also proposed is a bucketing method, where the embeddings from multiple layers of BERT are selected/merged to represent diverse linguistic aspects.  
We test our models with MS MARCO and TREC CAR, showing that our models outperforms other sparse models.
\end{abstract}

\section{Introduction}
Using neural models for representing and processing textual data has become a de-facto standard. Recent approaches to information retrieval (IR) and natural language processing (NLP) tasks adopt contextual language models (e.g., BERT \cite{devlin2019bert}), ameliorating both synonymy and polysemy problems associated with words \cite{zhan2020repbert,khattab2020colbert,ding2020rocketqa}. In these approaches, queries and documents are first encoded into contextual embeddings, either independently \cite{zhan2020repbert} or with interactions \cite{ding2020rocketqa}, resulting in low-dimensional dense representations. Then the documents are retrieved based on a similarity metric, such as cosine similarity, defined for two vectors. 

Despite the impressive results, the inherent properties of dense representations — low dimensional and dense — can pose a severe efficiency challenge for first-stage or full ranking. Since each dimension in a dense embedding is overloaded and entangled (i.e., polysemous) with the limited number of dimensions available, it is susceptible to false matches with large index sizes \cite{reimers2020curse}. Also, all the dimensions must participate in representing words, queries, and documents regardless of the amount of information content as well as in matching \cite{zamani2018neural}, which is inefficient. As a result, it is meaningless to build an inverted index for the dimensions, without which it is difficult to build an efficient and effective first-stage or full ranker. 

Other drawbacks of the dense embedding approaches to IR include: 1) The retrieval results are hardly interpretable like other neural approaches, making it difficult to improve the design through failure analyses or implement conditional/selective processing \cite{belinkov-glass-2019-analysis}; and 2) It is not straightforward to adopt the well studied term-based symbolic IR techniques, such as pseudo-relevance feedback, for further improvements. 

Our focus is to propose a full-blown neural first-stage ranker that alleviates the shortcomings of dense neural IR and yet achieves competitive effectiveness. The main thrust of our method is to utilize ultra-high dimensional (UHD) embeddings\footnote{The dimension size is close to half a million.} with high sparsity, possessing superior expressive and discriminating power to the extent that they are binarized. In addition, our proposed approach has the potential to make individual dimensions of the document/query embeddings interpretable \cite{faruqui2015sparse, sun2016sparse} and support mutually non-interfering aggregation of multiple embeddings \cite{ahmad2015properties}.

In order to obtain UHD embeddings, we train the $Winner\text{-}Take\text{-}All$ (WTA) model \cite{ahmad2015properties,makhzani2015winner} on top of BERT with a new learning objective for IR. WTA model is fundamentally a linear layer that only preserves top-$k$ activation and sets the others to zero. WTA is chosen because we can precisely control the outputs' sparsity by explicitly setting the parameter $k$. 
In contrast, the method relying on regularization for sparsity (e.g., minimizing L1-norm) \cite{faruqui2015sparse, sun2016sparse, zamani2018neural} is neither straightforward nor precise in controlling the degree of sparsity. 

With the large capacity and robustness against noise and interference, 
UHD allows for binarized representations so that matching can be further simplified with little degradation in effectiveness. This capability of UHD achieves extreme efficiency in terms of storage and speed, making it possible to build a stand-alone neural IR model for an industry.

Besides the efficiency goal, we also attempt to represent different aspects of linguistic properties of documents and queries. Instead of the common approach of using only the final layer of BERT, we make use of multiple layers, each of which emits token representations for different linguistic properties \cite{jo-myaeng-2020-roles}, by devising a bucketing mechanism.

We evaluate our model on MS MARCO passage retrieval \cite{bajaj2018ms} and TREC CAR \cite{dietz2017trec}, showing that it outperforms previous sparse models and is competitive with dense models for effectiveness. 
Even though it is a neural model, our UHD-based IR method with binarization is highly efficient, on par with BM25, surpassing all the sparse models.



    
    

\section{Related Work}
\label{sec::2_related_work}

State-of-the-art neural retrieval models \cite{nogueira2019passage,ding2020rocketqa} adopt a cross-encoder that shows high effectiveness while known to be impractical for large-scale retrieval \cite{luan2021sparse}. The cross-encoder requires that a query and documents must be encoded together, limiting models to a re-ranker. Dense neural models have been proposed for first-stage retrieval. RepBERT \cite{zhan2020repbert} encodes queries and documents separately and ranks using an inner product. It achieves efficiency by relying heavily on GPUs. ColBERT \cite{khattab2020colbert} positions itself between the cross-encoder and inner product by proposing a late interaction for scoring. Despite its promising performance, it remains questionable whether first-stage retrieval is computationally feasible at a large scale \cite{bai2020sparterm}, even with an external library Faiss \cite{8733051}.

Sparse models are attractive for first-stage retrieval \cite{10.1145/3397271.3401204, dai2020context, nogueira2019document, Nogueira2019FromDT, bai2020sparterm}.
Their retrieval speed comes from the sparsity that enables to leverage an efficient inverted index. On the other hand, they usually lack the ability to use non-symbolic latent semantics that can be captured by neural models. SOLAR \cite{Medini2020SOLARSO} proposes randomly initialized high-dimensional and ultra-sparse embeddings for book classification but ignores their content, making it unsuitable for IR tasks. SNRM \cite{zamani2018neural} creates a sparse latent representation using a fully connected layer similar to an autoencoder. It is limited with a simple word embedding, lower dimensionality (20K), uncontrollable sparsity.

\begin{figure*}[ht!]
  \centering
  \includegraphics[width=0.6\linewidth]{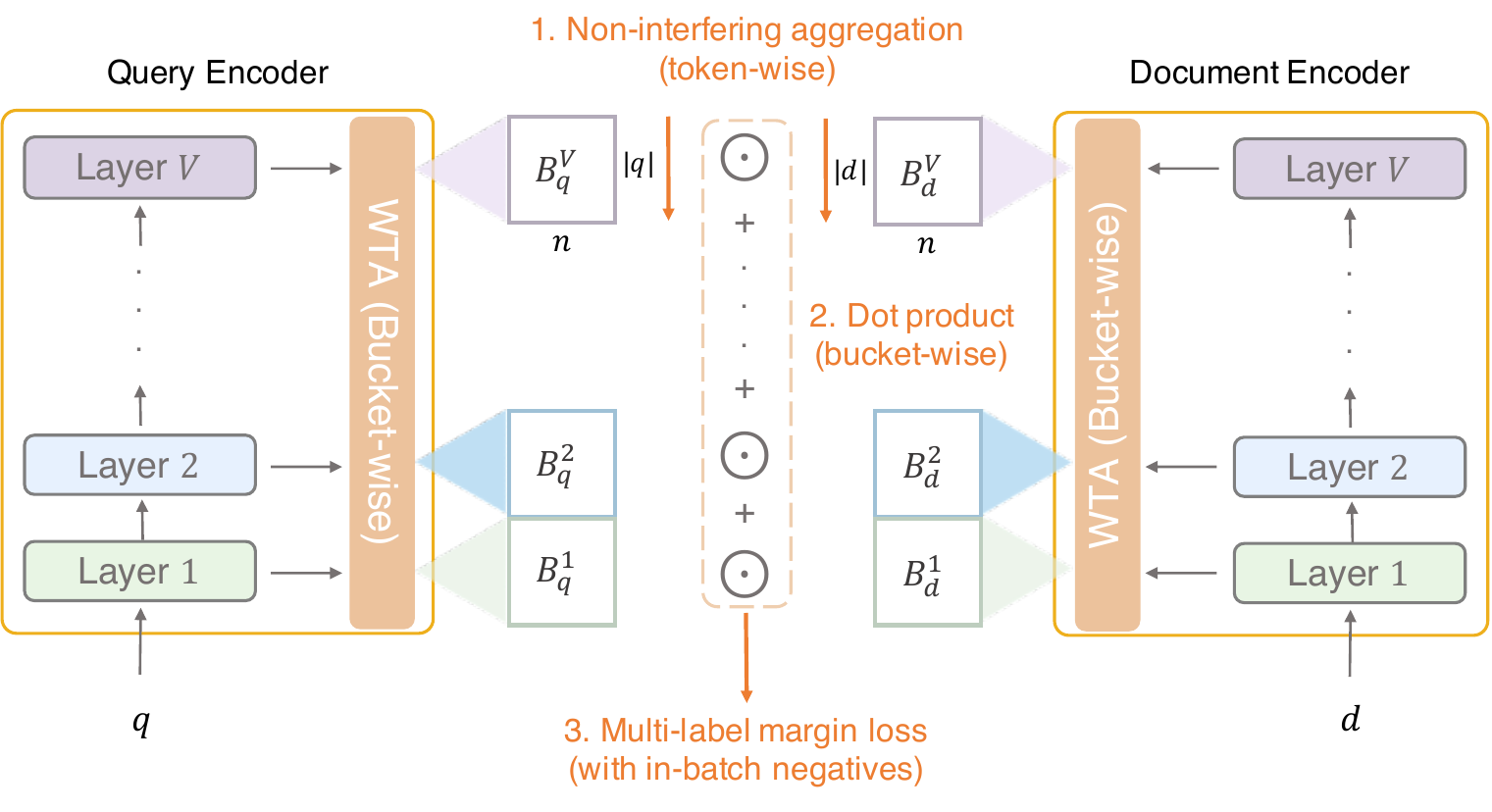}
  \caption{Overall architecture with the training scheme for building UHD sparse representations.}
  \label{fig:overall_architecture}
\end{figure*}

\section{Sparsified and Bucketized Embeddings} 

\subsection{Design Objectives}

Our goal is to build an efficient
neural ranker without relying on external efficiency-enhancing measures like Approximate Nearest Neighbor (ANN) \cite{8733051,jegou2010product}. We posit that it is crucial to represent the distinct textual signals in the embeddings for enhanced discriminative power, thereby achieving the effect of symbol-like characteristics with neural approaches.

Despite the contextualized language modeling capabilities of the recent transformer architectures, we note the shortcomings of the low-dimensional dense representations for IR: 1) All the dimensions must always be accessed during document-query matching; 2) Dimensions are highly overloaded and polysemous that they hardly serve as useful discriminators (e.g., \citealp{arora-etal-2018-linear,reimers2020curse}); 3) Various linguistic properties are entangled or under-represented.

As a result, we attempt to enforce the following:

\begin{itemize}
    \setlength\itemsep{0.0em}
    \item The embeddings need to be sufficiently high dimensional and sparse, so that they can be processed efficiently for matching while enjoying the high expressive power.
    
    \item Each dimension represents a specific concept, making it suitable for precise semantic computation and more interpretable.
    
    \item Different aspects of queries and documents should be captured for versatile representations as IR queries are notoriously ambiguous with diverse intents \cite{azad2019query}. 
\end{itemize}

Aside from representation perspectives, we adhere to the following requirements for efficiency:
\begin{itemize}
    \setlength\itemsep{0.0em}
    \item The ranker must support offline encoding of documents to build an inverted index.
    
    \item The matching function should ignore a signal if it is not strong enough and avoid unnecessary computation.
    
    
    \item Binarization should be possible for efficiency without unacceptable degradation of effectiveness \cite{ahmad2015properties, Zhou2020LearningBC}. 
\end{itemize}

\subsection{Overall Architecture}
Figure \ref{fig:overall_architecture} depicts the overall architecture of our Ultra-High Dimensional (UHD) model. Like most sparse models (e.g., \citealp{zamani2018neural,bai2020sparterm}), it comprises a query encoder, a document encoder, and a scoring function. While the query and document encoders are run separately, unlike interaction models (e.g., \citealp{ding2020rocketqa}), the weights are shared for the query and document encoders. After the final query and document representations are formed with sparsification and bucketization to be explained below, they are matched with dot product as the scoring function.

\begin{figure}[ht]
  \centering
  \includegraphics[width=0.8\linewidth]{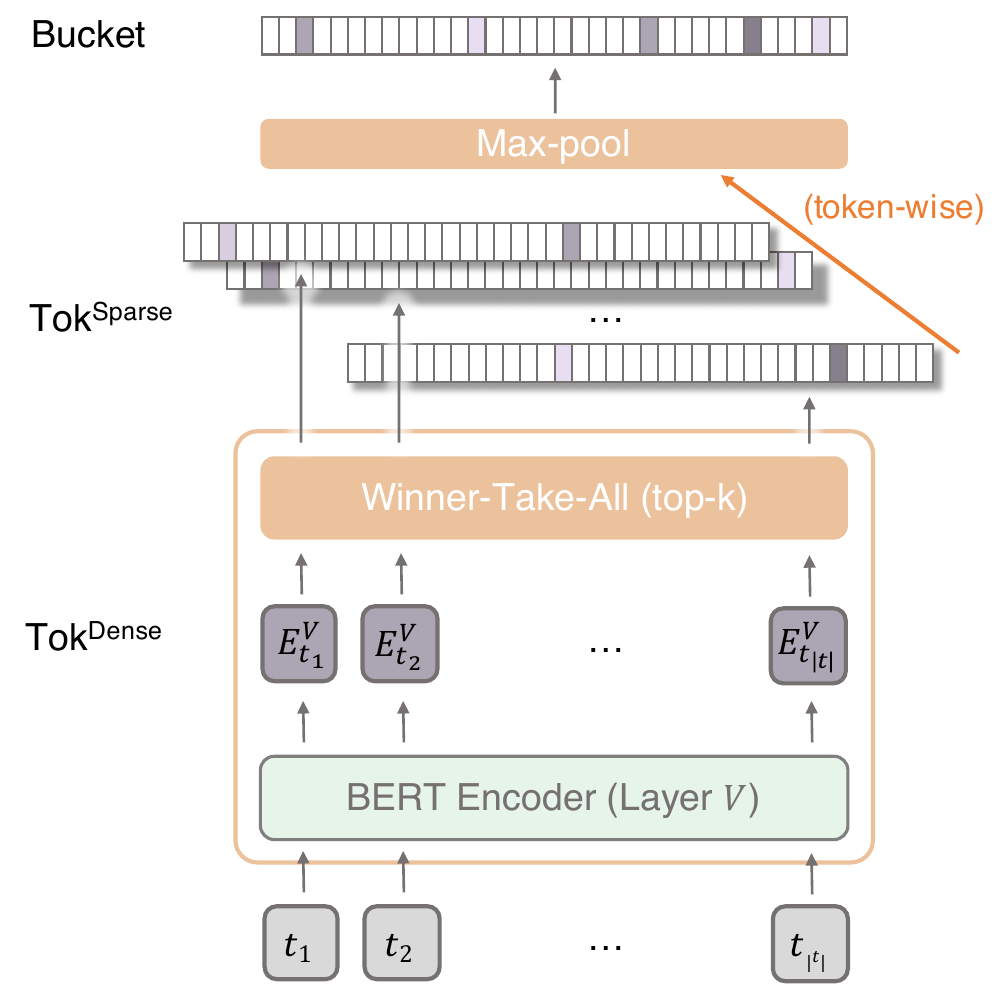}
  \caption{The Encoder structure.}
  \label{fig:encoder}
\end{figure}

The encoder is composed of three modules: 1) the BERT module to convert text into dense token embeddings, 2) the \textit{Winner\text{-}Take\text{-}All} (WTA) module that sparsifies the BERT module's outputs, and 3) the Max-pool module that performs \textit{non-interfering aggregation} of sparse token embeddings. We define the final output from the WTA module as a \textit{bucket} (Figure \ref{fig:encoder}), implying that the final representation to be used for matching contains multiple buckets. The BERT and WTA modules are trained jointly with an objective to maximize the similarity between a query bucket and individual relevant document buckets. 

Our UHD representations can be seen as bucketed with or segmented into multiple parts that represent different aspects of a query or document. From the procedural point of view, multiple buckets are produced for a query or document by different versions of WTA and concatenated to build the final UHD representation. 


\subsection{Query/Document Encoder}
\label{ssec:encoder}

Let $q = \{q_1, q_2, ..., q_{|q|}\}$ be a query and $d = \{d_1, d_2, ..., d_{|d|}\}$ be a document. We obtain contextualized dense representations from BERT for the tokens in $q$ and  $d$. Aside from the bucketed representations to be explained in Section \ref{sec::3_4_multibucket}, we here assume a common option of using only the last layer for a query or document representation. An embedding for $q$ or  $d$ is obtained as follows:

{\small
\begin{equation}
    \begin{aligned}
        E_q = BERT(q) \in \mathbb{R}^{|q| \times h}
        \\ 
        E_d = BERT(d) \in \mathbb{R}^{|d| \times h}
    \end{aligned}
    \label{eq:bert}
\end{equation}
}
where $h$ is the hidden size of BERT. Queries and documents are encoded separately with the same encoder.

To sparsify each dense token embedding, we adopt a  WTA layer \cite{makhzani2015winner}, which is an $n$-dimensional linear layer in which only top-$k$ activations ($k \ll n$) are preserved while the others are set to zero. 
Let $t_i$ be the $i^{th}$ token of a query or a document and $E_{t_i}$ be its dense embedding. Then a high-dimensional sparse representation $S_{t_i}$ is built as follows:

{\small
\begin{equation}
    z_{t_i} = E_{t_i} \cdot W + b,\ W \in \mathbb{R}^{h \times n},\ b \in \mathbb{R}^{n}
\end{equation}
\begin{equation}
    S_{t_i}[dim] = 
    \begin{cases}
      z_{t_i}[dim], & \text{if}\ z_{t_i}[dim] \in \text{top-}k(z_{t_i}) \\
      0, & \text{otherwise}
    \end{cases}
\end{equation}
}
where $dim$ $\in [1, n]$ is dimension of $z_{t_i}$ and $S_{t_i}$. 

In order to drive the learning flow to winning signals, the WTA module considers only the gradients flowing back through the \textit{winning} dimensions. In addition, we impose \textit{weight sparsity} constraint proposed in \citet{DBLP:journals/corr/abs-1903-11257}, which is like applying dropout \cite{JMLR:v15:srivastava14a} individually to output layer's nodes. The benefits of adopting WTA are: 1) We can control sparsity of a resulting embedding precisely and conveniently by adjusting $k$, an ability considered important for generating sparse representations reliably, and 2) $k$ can be modified at inference time so that the output's sparsity can be altered for an application's need without re-training the model.

Sparse token embeddings generated by the encoder are aggregated with token-wise max-pool followed by L2-normalization to produce a single sparse embedding $B_t$, a \textit{bucket}: 

{\small
\begin{equation}
    B_t = {max\text{-}pool}({S_{t_1}, S_{t_2}, ..., S_{t_{|t|}}}) \in \mathbb{R}^{n}
    \label{eq:bucket}
\end{equation}
}

Note that our sparse embeddings can be merged with minimal information interference because non-zero dimensions in high-dimensional space are not likely to overlap among the token embeddings, resulting in \textit{non-interfering} max-pooling. This effect is contrasted with dense representations that are prone to lose much information with aggregation. The ability of preserving token-level information, which is critical for IR \cite{khattab2020colbert}, makes our method storage efficient.

\subsection{Support for Binary Matching and Search}
\label{sec::3_4_binary}
With ultra-high dimensionality, it becomes possible to just count how many \textit{winning signals} of two representations overlap for similarity (binary matching); the probability they accidentally share the same winning signal becomes exponentially smaller with large dimension size \cite{ahmad2015properties,DBLP:journals/corr/abs-1903-11257}. With proper optimization (e.g., \cite{Zhou2020LearningBC,Tissier2019NearlosslessBO}), it becomes possible to perform highly efficient search while achieving comparable performance. In the experiment section, we empirically show that our model can support the binary matching with negligible impact on the effectiveness.


\subsection{Representations with Multiple Buckets}
\label{sec::3_4_multibucket}
We use multiple buckets to encode information with different levels of abstraction coming from different layers of BERT, as shown in \cite{jo-myaeng-2020-roles}. We expect UHD representations extracted from multiple layers contain richer information than from a single layer (e.g., the last layer often used for a downstream task). 


We first extract $V$ representations corresponding to the number of the BERT layers for all tokens $t$.

{\small
\begin{equation}
    E^j_t = BERT^j(t) \in \mathbb{R}^{|t| \times h} 
    \label{eq:vertical_bucket}
\end{equation}
}
where $j$ is a BERT layer $\in [1,\ V]$.
WTA layers are then independently applied to all BERT layers as in Section \ref{ssec:encoder} to obtain $V$ buckets. 

{\small
\begin{equation}
    B^j_t = WTA^j(E^j_t) \in \mathbb{R}^{n}
\end{equation}
}

After applying the bucketing mechanism, we obtain $B^{j}_q$ and $B^{j}_d$ for $q$ and $d$ so that a relevance score for the query and document is computed by a bucket-wise dot product as follows.

{\small
\begin{equation}
    \begin{aligned}
    Rel(q, d) = \sum_{j}B^{j}_q \cdot B^{j}_d
    \end{aligned}
    \label{eq:relevance_score}
\end{equation}
}

\subsection{Training}
\label{ssec:training}

The entire model is trained to make a query similar to the relevant documents and dissimilar to the irrelevant ones. Given a query $q$, a set of relevant (positive) documents $D^{p}$, and a set of irrelevant (negative) documents $D^{n}$, we calculate the loss: 

{\small
\begin{equation}
    \mathcal{L} =\sum_{\substack{d^p \in D^{p}\\ d^n \in D^{n}}} {\rm max}(0, 1-Rel(q, d^p)+Rel(q, d^n))
    \label{eq:loss}
\end{equation}
}
Given a query, we regard the positives of other queries within a mini-batch as negative samples (i.e., in-batch negatives), following \citet{zhan2020repbert}. 

\subsection{Ranking}


Our model allows for exploiting an inverted index.
We regard each dimension in our bucketed sparse representations, which is indexable, as a signal or a latent term. For instance, if $n$ (dimensionality) is 81,920, a document is represented with a combination of a few latent terms out of 81,920. Only the non-zero dimensions of a document enter the inverted index with their weights. The level of efficiency in symbolic IR can be achieved since only a small fraction of the dimension in our UHD representation contains a non-zero value. Even higher efficiency can be gained by using the binarized output for indexing and ranking. For binarization, we convert non-zero values to 1, leaving others as 0.

We first encode all documents in a collection using the trained encoder to construct an inverted index. Each bucket conceptually has its own independent inverted index, resulting in $|B|$ (e.g. the number of BERT layers) inverted indices. Note that this process is needed only once offline. At inference (retrieval) time, we encode a given query to make $|B|$ representations for bucket-wise dot products and sum up the scores for the final relevance score as in Eq. \ref{eq:relevance_score}. This bucket-wise index construction and scoring can be easily distributed for added efficiency.


\section{Experiments}
We conduct a series of experiments for validity of the proposed retrieval model and the associated methods against recently proposed sparse methods. We also juxtapose our results with those of the most recent dense models although they are not geared toward full ranking, without resorting to additional computational resources. We defer a qualitative analysis that shows UHD's interpretability to Appendix \ref{sec::a_4_interpret} due to the lack of space.

While SNRM \cite{zamani2018neural} would be a suitable baseline as a sparse model, we were unable to reproduce the model following the hyperparameter settings presented in the paper.
This reproducibility issue is also reported in \citet{Medini2020SOLARSO} and \citet{Paria2020Minimizing}.

Implementation details and hyperparameters are available in Appendix \ref{ssec::setting-details} for reproducibility.

\begingroup
\begin{table*}[ht]
\centering
\begin{tabularx}{\textwidth}{ Yssssmss }
\toprule
 & \multicolumn{4}{c}{\normalsize{\textbf{MS MARCO}}} & &  \multicolumn{2}{c}{\normalsize{\textbf{TREC CAR}}} \\
\normalsize{\textbf{Model}} & \small{\textbf{MRR@10}} & \small{\textbf{R@50}} & \small{\textbf{R@200}} & \small{\textbf{R@1000}} & & \small{\textbf{MRR@10}} & \small{\textbf{MAP}} \\ 
\midrule\midrule
\multicolumn{8}{c}{\normalsize{\textbf{Dense Embedding Approaches}}} \\
\midrule
RepBERT & 30.4 & - & - & 94 & & - & - \\
ColBERT & 36.0 & 82.9 & 92.3 & 96.8 & & 44.3$^{\dagger}$ & 31.3$^{\dagger}$ \\
BERT Base$^{\spadesuit}$ & 34.7$^{\dagger}$ & - & - & - & & - & 31.0$^{\dagger}$ \\
BERT Large$^{\spadesuit}$ & 36.5$^{\dagger}$ & - & - & - & & - & 33.5$^{\dagger}$ \\
\midrule
\multicolumn{8}{c}{\normalsize{\textbf{Sparse Representation Approaches}}} \\
\midrule
BM25 & 18.7 & 59.2 & 73.8 & 85.7 & & - & 15.3 \\
Doc2query & 21.5 & 64.4 & 77.9 & 89.1 & & - & 18.1 \\
DeepCT & 24.3 & 69 & 82 & 91 & & 33.2 & 24.6 \\
DocTTTTTquery & 27.7 & 75.6 & 86.9 & 94.7 & & - & - \\
SparTerm & 27.94 & 72.48 & 84.05 & 92.45 & & - & - \\
\textbf{UHD-BERT} & \textbf{30.04} & \textbf{77.77} & \textbf{88.81} & \textbf{96.01} & & \textbf{37.32}$^{\dagger}$ & \textbf{25.73}$^{\dagger}$ \\
\bottomrule
\end{tabularx}
\caption{Comparisons on MS MARCO and TREC CAR. The dense approaches suffering from longer latency are shown as a reference. 
$\spadesuit$ refers to full-interaction models used for re-ranking. $\dagger$ denotes re-ranking results after BM25 retrieval. Our model under TREC CAR employs re-ranking for comparability with the dense models. The baseline results are from the ColBERT paper, except for RepBERT and SparTerm which are from their own.
}
\label{table:overall_performance}
\end{table*}
\endgroup

\subsection{Settings}
\label{sec::4.1_setting}

\paragraph{Dataset} Following previous work, we evaluate on MS MARCO \cite{bajaj2018ms} Passage Retrieval and TREC Complex Answer Retrieval (CAR) \cite{dietz2017trec}. 

\textbf{MS MARCO}\footnote{Official dataset and evaluation scripts can be found in https://github.com/microsoft/MSMARCO-Passage-Ranking.} consists of 8.84M passages collected from the Web and 1M queries generated from real-world users of Bing. For training, we use 25\% of provided triples, (\textit{query}, \textit{positive passage}, \textit{negative passage}), in all our experiments unless otherwise specified. For evaluation, we use the dev set containing 6,980 queries. We mainly evaluate our model for full ranking, but in order to compare a large number of variants without an excessive computational burden, we also take advantage of the re-ranking setting (marked with ${\dagger}$). For evaluation metrics, we use MRR@10 and Recall@K.

\textbf{TREC CAR}\footnote{We used pre-processed dataset and evaluation scripts available here: https://github.com/nyu-dl/dl4marco-bert} is a synthetic dataset, consisting of approximately 29M Wikipedia passages and 3M queries. Following related work \cite{nogueira2019document,nogueira2019passage,khattab2020colbert}, we use the first four of five pre-defined folds as a training set, which consists of 5M query-passage pairs. For evaluation, we use the test set (2,254 queries) designated for an official run for TREC-CAR 2017. Our results are in MRR@10 and Mean Average Precision (MAP).

\paragraph{Baselines} We include both sparse methods for direct comparisons and dense methods as a reference, which require additional computational resources (e.g., GPU) for ranking.  

\textbf{BM25} \cite{robertson1994some} is a BoW-based sparse method with term weighting based on the query/document frequency signals.
  
\textbf{Doc2query} \cite{nogueira2019document} \& \textbf{DocTTTTTquery} \cite{Nogueira2019FromDT} is a sparse model that expands documents in the vocabulary space by predicting queries using BERT and T5 \cite{raffel2020exploring}, respectively. 

\textbf{DeepCT} \cite{10.1145/3397271.3401204} is a contextualized term weighting framework, which maps BERT representations to context-aware term weights. 

\textbf{SparTerm} \cite{bai2020sparterm} is a sparse model that generates term-weighted \& term-expanded sparse vectors belonging to the vocabulary space. 
 
\textbf{RepBERT} \cite{zhan2020repbert} is a dense model that encodes queries and documents separately, with mean-pooling on BERT token embeddings.

\textbf{ColBERT} \cite{khattab2020colbert} is a dense model with token-level late interactions between queries and passages, which are encoded separately.

\textbf{BERT Base \& Large} \cite{nogueira2019passage} are dense models that fully leverage the interaction between a query and passage. They are often used to re-rank fast BM25 results.

\subsection{Overall Effectiveness}

Table \ref{table:overall_performance} shows the full-ranking (except for $\dagger$\footnote{Re-ranking scores tend to be worse than full-ranking as shown in \citet{khattab2020colbert}}) performance of the proposed model (UHD-BERT) against the baselines on the two datasets. 
Among the sparse approaches, UHD-BERT outperforms all the baseline models, from traditional term-based methods to recent neural approaches. 
Unlike the previous sparse approaches that mainly focus on efficiency at the expense of losing information, our approach achieves richer embeddings using bucketed UHD sparse representations while maintaining the necessary sparsity for efficiency. For this result, we train UHD-BERT with buckets on the layers \{$2,4,6,8,10,12$\}, with the dimensionality $n$ being 81,920 and non-zero dimensions $k$ being 80.

The performance of the dense models is also provided in Table \ref{table:overall_performance}. Since they are not suitable for full-ranking without additional efficiency-enhancing measures, the result is only a reference. As expected, they show higher effectiveness than UHD-BERT, due to their heavy interactions between the query and document representations, requiring an inference time overhead. While ColBERT \cite{khattab2020colbert} is most promising with the later interaction idea, it still requires relatively heavy computation, which might be impractical for industrial applications. RepBERT \cite{zhan2020repbert} needs to employ external resources (i.e., 
GPUs) for comparable efficiency. UHD-BERT is on par with RepBERT without such overheads, highlighting the advantage of our model being as efficient as the conventional symbolic IR models and yet approaching to the effectiveness of the dense models requiring heavy interactions. 

\subsection{Efficiency and Binarized UHD}
\label{sec::4.3_analysis on efficiency}

\begin{table}
\begin{center}
\begin{tabular}{lc}
\toprule
\textbf{Model} & \textbf{Complexity}   \\ \midrule 
RepBERT & $O(|C| * h)$ \\ 
ColBERT & $O(|C| * |q| * |d| * h)$ \\
\hline
UHD-BERT  & $O(|C| * k_q * k_d  / n)$ \\ 
\bottomrule
\end{tabular}
\end{center}
\caption{Complexity comparison between UHD-BERT and two dense baselines. $h$ denotes BERT dimension size. $n(\gg h)$ and $k$ denote the total and non-zero UHD-BERT dimension sizes, respectively. $|C|$ denotes the collection size. Increasing the dimension size enhances efficiency in UHD-BERT unlike dense models.}
\label{table:complex}
\end{table}

Since UHD-BERT is designed for full ranking, efficiency is as important as its effectiveness. Due to its sparse nature, efficiency of UHD depends on two factors, $n$ (dimensionality) and $k$ (the number of non-zero dimensions). For retrieval, only the non-zero dimensions of a query are used to search the inverted index. Therefore, the time complexity of the retrieval process is:

\begin{equation}
    O(k_q * d_{active}) = O(k_q * k_d / n)
\end{equation}

where $k_q$ and $k_d$ are $k$ for queries and documents, respectively. The value of $d_{active}$, the average number of  linked documents per dimension, is approximated by the probability that a dimension becomes non-zero, $k_d / n$. 

Table \ref{table:complex} shows that UHD-BERT has a much lower time complexity than the two dense models, ColBERT and RepBERT, as the actual ratio between the product of $k_q$ and $k_d$ and $n$ is less than $h$. In fact, the complexity decreases as the dimension size $n$ increases when $k$ is fixed to a constant for limited computing resources. Note, however, that $n$ is also resource-dependent and can account for a trade-off between efficiency and effectiveness together with $k$ as in Section \ref{sec::4.5_dimensionality}.

\begin{figure}[ht]
  \centering
  \includegraphics[width=0.9\linewidth]{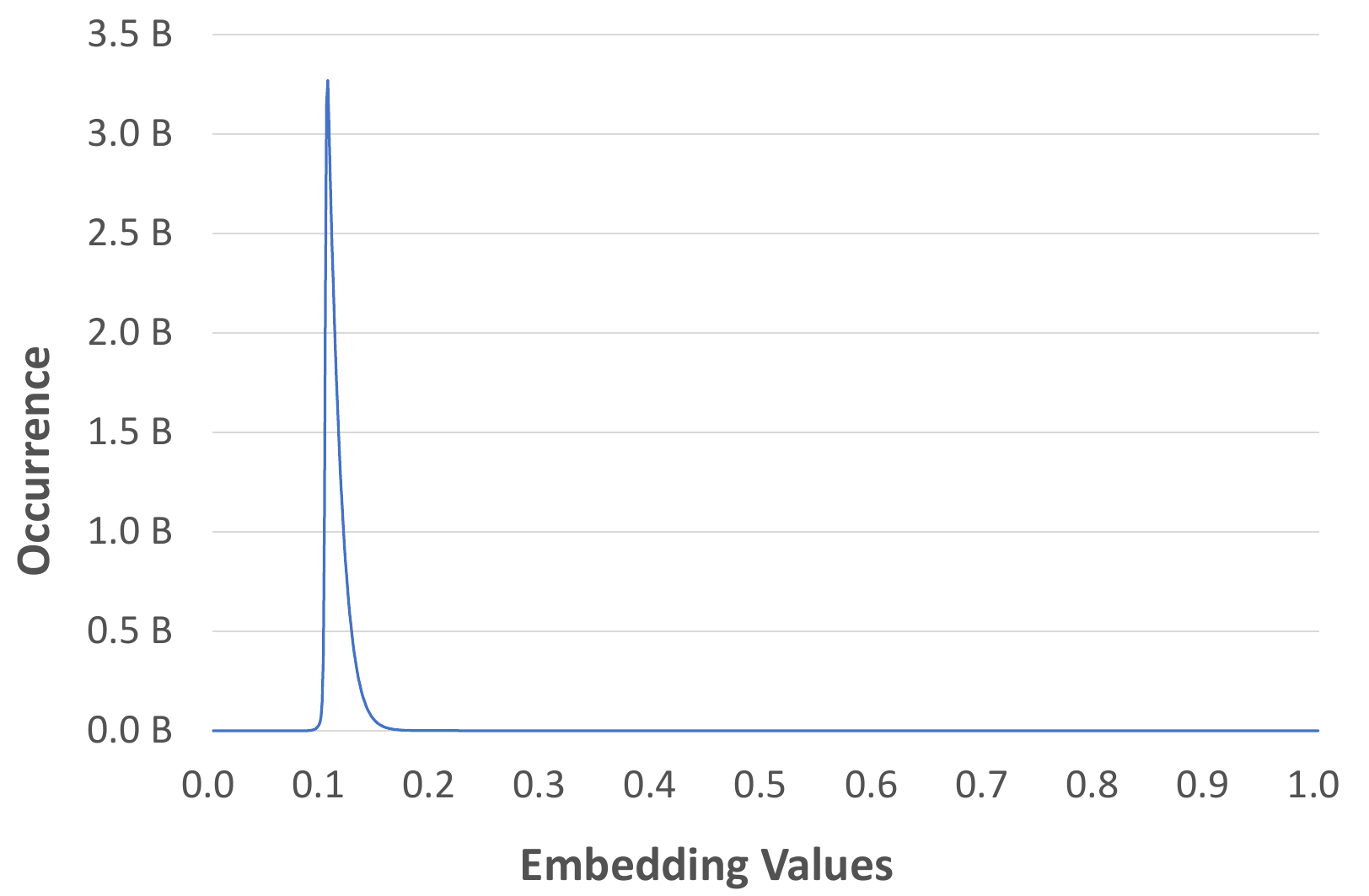}
  \caption{Distribution of embedding values on UHD.}
  \label{fig:score_distribtuion}
\end{figure}

For further improvement in efficiency, we experiment with a binarized UHD-BERT, feasibility of which is justified with the distribution of embedding values. It turns out that the embedding values are clustered into the range between 0.1 and 0.2 as in Figure \ref{fig:score_distribtuion} with the mean and standard deviation being 0.11169 and 0.00182, respectively. This unique pattern assures that binarization can be done without having to coerce the embedding values to either 0 or 1 unnaturally. With binarized UHD embeddings, the retrieval effectiveness remains almost identical (30.03) to the original (30.04) (Table \ref{table:overall_performance}). 

\begin{table}
\begin{center}
\begin{tabular}{lc}
\toprule
\textbf{Model} & \textbf{Latency(ms)}   \\ \midrule 
RepBERT & 80$^{\dagger}$ \\ 
ColBERT & 458$^{\dagger}$ \\ 
BERT Base & - \\ 
BERT Large & 3,400$^{\dagger}$ \\ 
\hline
BM25 & 62 \\ 
Doc2query & 85 \\ 
DeepCT & 62 \it{(est.)} \\ 
DocTTTTTquery & 87 \\ 
SparTerm & - \\ 
\hline
UHD Binarized Inverted Index & 63 \\ 
\bottomrule
\end{tabular}
\end{center}
\caption{Comparison of latency (ms/query) between our UHD binarized inverted index and the baseline models on the MS MARCO dev set. Baseline results are from the ColBERT paper, except for the RepBERT value from its own paper. ${\dagger}$ denotes GPU-accelerated document ranking.}
\label{table:latency}
\end{table}

As a result, our experiment on latency is based on an inverted index with a binarized version of UHD-BERT\footnote{The single bucket option was used but multiple buckets can be easily introduced with server-level parallelism.}. Table \ref{table:latency} shows comparisons against the baselines for latency, which in our case includes query preprocessing and encoding, and document scoring and sorting. The result clearly shows the efficiency advantage of using the inverted index with UHD-BERT over the baselines; it is almost identical to BM25, known to be extremely efficient.

\subsection{Analysis on Dimensionality and Sparsity}
\label{sec::4.5_dimensionality}

\begin{figure}[ht]
  \centering
  \includegraphics[width=\linewidth]{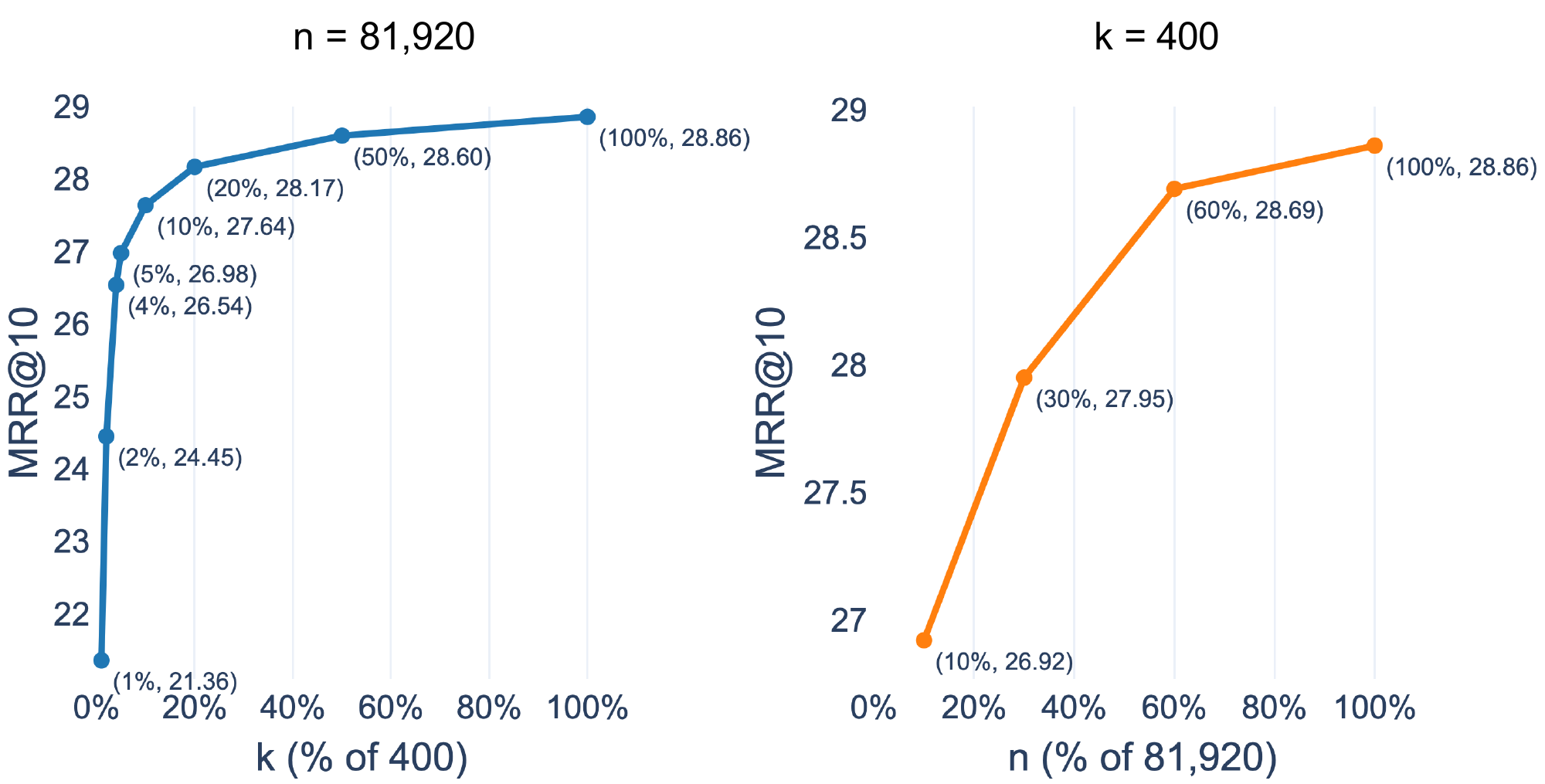}
  \caption{Impact of $n$ (dimensionality) and $k$ (non-zero dimensions).}
  \label{fig:NK}
\end{figure}

To understand the roles of $n$ (dimensionality) and $k$ (non-zero dimensions), we test our model with two different settings. Figure \ref{fig:NK} shows the performance trends: the impact of varying $k$ with \{4, 8, 16, 20, 40, 80, 200, 400\} on the left using the fixed $n=81,920$ and the impact of varying $n$ with \{8,192, 24,576, 49,152, 81,920\} on the right using the fixed $k= 400$. 

Given the fixed $n$, it is evident that the higher $k$, the better the score. This is because the absolute amount of information for representing a query and a document increases. The graph shows, however, that the score rises rapidly to a point and almost stays. It is not desirable to set an exorbitant $k$ (e.g., 8,192) as it would yield high computational costs for a very small gain. We see the score is reasonable even with $k$ being 16 (4\%). This suggests that is is important to find an optimal $k$ that satisfies the trade-off between effectiveness and efficiency.

Next, we observe that the score improves as the $n$ increases even with the fixed $k$. This is because the number of available activation patterns (i.e., expressive power) increases exponentially with larger $n$. Also, note that larger $n$ with the same $k$ means the increased sparsity (i.e., efficiency). This observation supports the motivation of UHD representations that endow the discrimination power while increasing the efficiency.

\begin{table}[h]
\begin{tabularx}{\linewidth}{Y Y Y Y}
    \toprule
     \textbf{Query $k$} & \textbf{MS MARCO} & \textbf{TREC CAR} \\
     \hline
     50 & 29.50 & 34.10 \\ 
     100 & \textbf{30.04} & 35.43 \\ 
     200 & 29.86 & 36.31 \\ 
     300 & 29.64 & \textbf{37.32} \\ 
     400 & 29.29 & 37.22 \\ 
     \hline
     \small{Original$\approx$1,200} & 28.99 & 37.12 \\ 
    \bottomrule
\end{tabularx} 
\caption{MRR@10 with different query $k$ (\# of non-zero dimensions of query after max-pool).}
\label{table:symb_query_k}
\end{table}

Finally, in order to analyze how query's sparsity affects the performance, 
we constrain each query's final $k$ (query $k$) after the max-pool operation and measure MRR@10, expecting a removal of trivial information that can cause query-drift. Table \ref{table:symb_query_k} shows how MRR@10 changes as we change the query $k$. The highest performance is achieved with $k$ being much lower than the original ($\approx$1,200). This means removing trivial information indeed helps improve matching quality. Another huge benefit is that it reduces the amount of required computation for matching.

\subsection{Analysis on Multiple Buckets}
\label{sec::4.6_multiBucket}

\begin{table}[h]
\begin{center}
\begin{tabular}{lc}
\toprule
\textbf{Strategy} & \textbf{MRR@10}   \\ \midrule 
Single-Bucket \\ ($b=1$, $n=49,152$, $k=480$) & 26.18 \\ 
\hline
Multi-Bucket \\ ($b=6$, $n=8,192$, $k=80$) & 27.08  \\ 
\hspace{0.15cm}  + bucket weight tuning & \textbf{27.41} \\
\bottomrule
\end{tabular}
\end{center}
\caption{Comparison of bucketing strategies on the MS MARCO dev set re-ranking task. The total dimensionality ($n$) and non-zero dimension size ($k$) are the same for both strategies. $b$ stands for the number of buckets.}
\label{table:multilayer}
\end{table}

In order to verify the effectiveness of our bucketing strategy, we compare the performance of single and multi-bucket strategies in Table \ref{table:multilayer}.
For a fair comparison, we set the total dimensionality ($n$) and the number of non-zero dimensions ($k$) to be the same for both of them. 
For the mutliple bucket setting, we create each bucket with the 2nd, 4th, 6th, 8th, 10th, and 12th layers of BERT, whereas the 12th layer of BERT is used for the single-bucket setting. For training, we only use 3\% of the MS MARCO Train Triples Small set. 

Our multi-bucket strategy is shown to give a good improvement, and the bucket weighting scheme improves it further. This result validates the idea explained in Section \ref{sec::3_4_multibucket} that it is crucial for IR tasks to exploit low- and mid-level lexical and syntactic information, as well as the last layer containing well-refined semantics. We provide additional analyses on other settings in Appendix \ref{sec::a4_buckets}.

\section{Conclusion}
We present UHD-BERT, a novel retrieval method empowered by extremely high dimensionality and sparsity that is easily controllable. To fully utilize the capacity of our representation scheme, we propose a bucketing mechanism that incorporates different linguistic aspects extracted from BERT and achieves the goal of building a neural retrieval model that ameliorates the synonymy and polysemy problems of symbolic text retrieval methods.
With binarization of the learned UHD representations and resulting inverted indices, we attain the desired efficiency that is comparable with BM25 and hence suitable for industrial applications. The benefits of the UHD representations and the resulting retrieval model were demonstrated with two different IR datasets, in comparison with the recent approaches to sparse representation methods. We also show that our method has a significant advantage in efficiency over the state-of-art dense models, with a slightly lower but competitive effectiveness. We plan to investigate how the query-document interactions and symbolic IR techniques can be incorporated for further improvements.

\section*{Acknowledgements}
This work was supported by the joint research program: "Implementing and Evaluating the Neural Matching Method for Efficient and Effective Search Engine" funded by Naver corporation. This work was also supported by Institute for Information \& communications Technology Planning \& Evaluation(IITP) grant funded by the Korea government(MSIT) (No. 2013-2-00131, Development of Knowledge Evolutionary WiseQA Platform Technology for Human Knowledge Augmented Services).



\bibliography{anthology,custom}

\begin{thebibliography}{35}
\expandafter\ifx\csname natexlab\endcsname\relax\def\natexlab#1{#1}\fi

\bibitem[{Ahmad and Hawkins(2015)}]{ahmad2015properties}
Subutai Ahmad and Jeff Hawkins. 2015.
\newblock Properties of sparse distributed representations and their
  application to hierarchical temporal memory.
\newblock \emph{arXiv preprint arXiv:1503.07469}.

\bibitem[{Ahmad and Scheinkman(2019)}]{DBLP:journals/corr/abs-1903-11257}
Subutai Ahmad and Luiz Scheinkman. 2019.
\newblock \href {http://arxiv.org/abs/1903.11257} {How can we be so dense? the
  benefits of using highly sparse representations}.
\newblock \emph{CoRR}, abs/1903.11257.

\bibitem[{Arora et~al.(2018)Arora, Li, Liang, Ma, and
  Risteski}]{arora-etal-2018-linear}
Sanjeev Arora, Yuanzhi Li, Yingyu Liang, Tengyu Ma, and Andrej Risteski. 2018.
\newblock \href {https://doi.org/10.1162/tacl_a_00034} {Linear algebraic
  structure of word senses, with applications to polysemy}.
\newblock \emph{Transactions of the Association for Computational Linguistics},
  6:483--495.

\bibitem[{Azad and Deepak(2019)}]{azad2019query}
Hiteshwar~Kumar Azad and Akshay Deepak. 2019.
\newblock Query expansion techniques for information retrieval: a survey.
\newblock \emph{Information Processing \& Management}, 56(5):1698--1735.

\bibitem[{Bai et~al.(2020)Bai, Li, Wang, Zhang, Shang, Xu, Wang, Wang, and
  Liu}]{bai2020sparterm}
Yang Bai, Xiaoguang Li, Gang Wang, Chaoliang Zhang, Lifeng Shang, Jun Xu,
  Zhaowei Wang, Fangshan Wang, and Qun Liu. 2020.
\newblock Sparterm: Learning term-based sparse representation for fast text
  retrieval.
\newblock \emph{arXiv preprint arXiv:2010.00768}.

\bibitem[{Bajaj et~al.(2018)Bajaj, Campos, Craswell, Deng, Gao, Liu, Majumder,
  McNamara, Mitra, Nguyen, Rosenberg, Song, Stoica, Tiwary, and
  Wang}]{bajaj2018ms}
Payal Bajaj, Daniel Campos, Nick Craswell, Li~Deng, Jianfeng Gao, Xiaodong Liu,
  Rangan Majumder, Andrew McNamara, Bhaskar Mitra, Tri Nguyen, Mir Rosenberg,
  Xia Song, Alina Stoica, Saurabh Tiwary, and Tong Wang. 2018.
\newblock \href {http://arxiv.org/abs/1611.09268} {Ms marco: A human generated
  machine reading comprehension dataset}.

\bibitem[{Belinkov and Glass(2019)}]{belinkov-glass-2019-analysis}
Yonatan Belinkov and James Glass. 2019.
\newblock \href {https://doi.org/10.1162/tacl_a_00254} {Analysis methods in
  neural language processing: A survey}.
\newblock \emph{Transactions of the Association for Computational Linguistics},
  7:49--72.

\bibitem[{Dai and Callan(2020{\natexlab{a}})}]{dai2020context}
Zhuyun Dai and Jamie Callan. 2020{\natexlab{a}}.
\newblock \href {https://doi.org/10.1145/3366423.3380258} {Context-aware
  document term weighting for ad-hoc search}.
\newblock In \emph{{WWW} '20: The Web Conference 2020, Taipei, Taiwan, April
  20-24, 2020}, pages 1897--1907. {ACM} / {IW3C2}.

\bibitem[{Dai and Callan(2020{\natexlab{b}})}]{10.1145/3397271.3401204}
Zhuyun Dai and Jamie Callan. 2020{\natexlab{b}}.
\newblock \href {https://doi.org/10.1145/3397271.3401204} {Context-aware term
  weighting for first stage passage retrieval}.
\newblock In \emph{Proceedings of the 43rd International {ACM} {SIGIR}
  conference on research and development in Information Retrieval, {SIGIR}
  2020, Virtual Event, China, July 25-30, 2020}, pages 1533--1536. {ACM}.

\bibitem[{Devlin et~al.(2019)Devlin, Chang, Lee, and
  Toutanova}]{devlin2019bert}
Jacob Devlin, Ming-Wei Chang, Kenton Lee, and Kristina Toutanova. 2019.
\newblock \href {https://doi.org/10.18653/v1/N19-1423} {{BERT}: Pre-training of
  deep bidirectional transformers for language understanding}.
\newblock In \emph{Proceedings of the 2019 Conference of the North {A}merican
  Chapter of the Association for Computational Linguistics: Human Language
  Technologies, Volume 1 (Long and Short Papers)}, pages 4171--4186,
  Minneapolis, Minnesota. Association for Computational Linguistics.

\bibitem[{Dietz et~al.(2017)Dietz, Verma, Radlinski, and
  Craswell}]{dietz2017trec}
Laura Dietz, Manisha Verma, Filip Radlinski, and Nick Craswell. 2017.
\newblock Trec complex answer retrieval overview.
\newblock TREC.

\bibitem[{Faruqui et~al.(2015)Faruqui, Tsvetkov, Yogatama, Dyer, and
  Smith}]{faruqui2015sparse}
Manaal Faruqui, Yulia Tsvetkov, Dani Yogatama, Chris Dyer, and Noah~A. Smith.
  2015.
\newblock \href {https://doi.org/10.3115/v1/P15-1144} {Sparse overcomplete word
  vector representations}.
\newblock In \emph{Proceedings of the 53rd Annual Meeting of the Association
  for Computational Linguistics and the 7th International Joint Conference on
  Natural Language Processing (Volume 1: Long Papers)}, pages 1491--1500,
  Beijing, China. Association for Computational Linguistics.

\bibitem[{Hendrycks and Gimpel(2017)}]{hendrycks2016bridging}
Dan Hendrycks and Kevin Gimpel. 2017.
\newblock \href {https://openreview.net/forum?id=Hkg4TI9xl} {A baseline for
  detecting misclassified and out-of-distribution examples in neural networks}.
\newblock In \emph{5th International Conference on Learning Representations,
  {ICLR} 2017, Toulon, France, April 24-26, 2017, Conference Track
  Proceedings}. OpenReview.net.

\bibitem[{Jegou et~al.(2010)Jegou, Douze, and Schmid}]{jegou2010product}
Herve Jegou, Matthijs Douze, and Cordelia Schmid. 2010.
\newblock Product quantization for nearest neighbor search.
\newblock \emph{IEEE transactions on pattern analysis and machine
  intelligence}, 33(1):117--128.

\bibitem[{Jo and Myaeng(2020)}]{jo-myaeng-2020-roles}
Jae-young Jo and Sung-Hyon Myaeng. 2020.
\newblock \href {https://doi.org/10.18653/v1/2020.acl-main.311} {Roles and
  utilization of attention heads in transformer-based neural language models}.
\newblock In \emph{Proceedings of the 58th Annual Meeting of the Association
  for Computational Linguistics}, pages 3404--3417, Online. Association for
  Computational Linguistics.

\bibitem[{Johnson et~al.(2019)Johnson, Douze, and J{\'e}gou}]{8733051}
Jeff Johnson, Matthijs Douze, and Herv{\'e} J{\'e}gou. 2019.
\newblock Billion-scale similarity search with gpus.
\newblock \emph{IEEE Transactions on Big Data}.

\bibitem[{Khattab and Zaharia(2020)}]{khattab2020colbert}
Omar Khattab and Matei Zaharia. 2020.
\newblock \href {https://doi.org/10.1145/3397271.3401075} {Colbert: Efficient
  and effective passage search via contextualized late interaction over
  {BERT}}.
\newblock In \emph{Proceedings of the 43rd International {ACM} {SIGIR}
  conference on research and development in Information Retrieval, {SIGIR}
  2020, Virtual Event, China, July 25-30, 2020}, pages 39--48. {ACM}.

\bibitem[{Luan et~al.(2021)Luan, Eisenstein, Toutanova, and
  Collins}]{luan2021sparse}
Yi~Luan, Jacob Eisenstein, Kristina Toutanova, and Michael Collins. 2021.
\newblock Sparse, dense, and attentional representations for text retrieval.
\newblock \emph{Transactions of the Association for Computational Linguistics},
  9:329--345.

\bibitem[{Makhzani and Frey(2015)}]{makhzani2015winner}
Alireza Makhzani and Brendan~J. Frey. 2015.
\newblock \href
  {https://proceedings.neurips.cc/paper/2015/hash/5129a5ddcd0dcd755232baa04c231698-Abstract.html}
  {Winner-take-all autoencoders}.
\newblock In \emph{Advances in Neural Information Processing Systems 28: Annual
  Conference on Neural Information Processing Systems 2015, December 7-12,
  2015, Montreal, Quebec, Canada}, pages 2791--2799.

\bibitem[{Medini et~al.(2020)Medini, di~Chen, and
  Shrivastava}]{Medini2020SOLARSO}
Tharun Medini, Bei di~Chen, and Anshumali Shrivastava. 2020.
\newblock Solar: Sparse orthogonal learned and random embeddings.
\newblock \emph{ArXiv}, abs/2008.13225.

\bibitem[{Nogueira(2019)}]{Nogueira2019FromDT}
Rodrigo Nogueira. 2019.
\newblock From doc2query to doctttttquery.

\bibitem[{Nogueira and Cho(2019)}]{nogueira2019passage}
Rodrigo Nogueira and Kyunghyun Cho. 2019.
\newblock Passage re-ranking with bert.
\newblock \emph{arXiv preprint arXiv:1901.04085}.

\bibitem[{Nogueira et~al.(2019)Nogueira, Yang, Lin, and
  Cho}]{nogueira2019document}
Rodrigo Nogueira, Wei Yang, Jimmy Lin, and Kyunghyun Cho. 2019.
\newblock Document expansion by query prediction.
\newblock \emph{arXiv preprint arXiv:1904.08375}.

\bibitem[{Paria et~al.(2020)Paria, Yeh, Yen, Xu, Ravikumar, and
  P{\'{o}}czos}]{Paria2020Minimizing}
Biswajit Paria, Chih{-}Kuan Yeh, Ian~En{-}Hsu Yen, Ning Xu, Pradeep Ravikumar,
  and Barnab{\'{a}}s P{\'{o}}czos. 2020.
\newblock \href {https://openreview.net/forum?id=SygpC6Ntvr} {Minimizing flops
  to learn efficient sparse representations}.
\newblock In \emph{8th International Conference on Learning Representations,
  {ICLR} 2020, Addis Ababa, Ethiopia, April 26-30, 2020}. OpenReview.net.

\bibitem[{Qu et~al.(2021)Qu, Ding, Liu, Liu, Ren, Zhao, Dong, Wu, and
  Wang}]{ding2020rocketqa}
Yingqi Qu, Yuchen Ding, Jing Liu, Kai Liu, Ruiyang Ren, Wayne~Xin Zhao, Daxiang
  Dong, Hua Wu, and Haifeng Wang. 2021.
\newblock \href {https://www.aclweb.org/anthology/2021.naacl-main.466}
  {{R}ocket{QA}: An optimized training approach to dense passage retrieval for
  open-domain question answering}.
\newblock In \emph{Proceedings of the 2021 Conference of the North American
  Chapter of the Association for Computational Linguistics: Human Language
  Technologies}, pages 5835--5847, Online. Association for Computational
  Linguistics.

\bibitem[{Raffel et~al.(2020)Raffel, Shazeer, Roberts, Lee, Narang, Matena,
  Zhou, Li, and Liu}]{raffel2020exploring}
Colin Raffel, Noam Shazeer, Adam Roberts, Katherine Lee, Sharan Narang, Michael
  Matena, Yanqi Zhou, Wei Li, and Peter~J Liu. 2020.
\newblock Exploring the limits of transfer learning with a unified text-to-text
  transformer.
\newblock \emph{Journal of Machine Learning Research}, 21:1--67.

\bibitem[{Reimers and Gurevych(2020)}]{reimers2020curse}
Nils Reimers and Iryna Gurevych. 2020.
\newblock The curse of dense low-dimensional information retrieval for large
  index sizes.
\newblock \emph{arXiv preprint arXiv:2012.14210}.

\bibitem[{Robertson and Walker(1994)}]{robertson1994some}
Stephen~E Robertson and Steve Walker. 1994.
\newblock Some simple effective approximations to the 2-poisson model for
  probabilistic weighted retrieval.
\newblock In \emph{SIGIR’94}, pages 232--241. Springer.

\bibitem[{Srivastava et~al.(2014)Srivastava, Hinton, Krizhevsky, Sutskever, and
  Salakhutdinov}]{JMLR:v15:srivastava14a}
Nitish Srivastava, Geoffrey Hinton, Alex Krizhevsky, Ilya Sutskever, and Ruslan
  Salakhutdinov. 2014.
\newblock \href {http://jmlr.org/papers/v15/srivastava14a.html} {Dropout: A
  simple way to prevent neural networks from overfitting}.
\newblock \emph{Journal of Machine Learning Research}, 15(56):1929--1958.

\bibitem[{Sun et~al.(2016)Sun, Guo, Lan, Xu, and Cheng}]{sun2016sparse}
Fei Sun, Jiafeng Guo, Yanyan Lan, Jun Xu, and Xueqi Cheng. 2016.
\newblock Sparse word embeddings using l1 regularized online learning.
\newblock In \emph{Proceedings of the Twenty-Fifth International Joint
  Conference on Artificial Intelligence}, pages 2915--2921.

\bibitem[{Tissier et~al.(2019)Tissier, Gravier, and
  Habrard}]{Tissier2019NearlosslessBO}
Julien Tissier, Christophe Gravier, and Amaury Habrard. 2019.
\newblock Near-lossless binarization of word embeddings.
\newblock In \emph{Proceedings of the AAAI Conference on Artificial
  Intelligence}, volume~33, pages 7104--7111.

\bibitem[{Wolf et~al.(2020)Wolf, Debut, Sanh, Chaumond, Delangue, Moi, Cistac,
  Rault, Louf, Funtowicz, Davison, Shleifer, von Platen, Ma, Jernite, Plu, Xu,
  Le~Scao, Gugger, Drame, Lhoest, and Rush}]{wolf-etal-2020-transformers}
Thomas Wolf, Lysandre Debut, Victor Sanh, Julien Chaumond, Clement Delangue,
  Anthony Moi, Pierric Cistac, Tim Rault, Remi Louf, Morgan Funtowicz, Joe
  Davison, Sam Shleifer, Patrick von Platen, Clara Ma, Yacine Jernite, Julien
  Plu, Canwen Xu, Teven Le~Scao, Sylvain Gugger, Mariama Drame, Quentin Lhoest,
  and Alexander Rush. 2020.
\newblock \href {https://doi.org/10.18653/v1/2020.emnlp-demos.6} {Transformers:
  State-of-the-art natural language processing}.
\newblock In \emph{Proceedings of the 2020 Conference on Empirical Methods in
  Natural Language Processing: System Demonstrations}, pages 38--45, Online.
  Association for Computational Linguistics.

\bibitem[{Zamani et~al.(2018)Zamani, Dehghani, Croft, Learned{-}Miller, and
  Kamps}]{zamani2018neural}
Hamed Zamani, Mostafa Dehghani, W.~Bruce Croft, Erik~G. Learned{-}Miller, and
  Jaap Kamps. 2018.
\newblock \href {https://doi.org/10.1145/3269206.3271800} {From neural
  re-ranking to neural ranking: Learning a sparse representation for inverted
  indexing}.
\newblock In \emph{Proceedings of the 27th {ACM} International Conference on
  Information and Knowledge Management, {CIKM} 2018, Torino, Italy, October
  22-26, 2018}, pages 497--506. {ACM}.

\bibitem[{Zhan et~al.(2020)Zhan, Mao, Liu, Zhang, and Ma}]{zhan2020repbert}
Jingtao Zhan, Jiaxin Mao, Yiqun Liu, Min Zhang, and Shaoping Ma. 2020.
\newblock Repbert: Contextualized text embeddings for first-stage retrieval.
\newblock \emph{arXiv preprint arXiv:2006.15498}.

\bibitem[{Zhou et~al.(2020)Zhou, Bai, Liu, Zhou, and
  Hancock}]{Zhou2020LearningBC}
Lei Zhou, Xiao Bai, Xianglong Liu, Jun Zhou, and Edwin~R Hancock. 2020.
\newblock Learning binary code for fast nearest subspace search.
\newblock \emph{Pattern Recognition}, 98:107040.

\end{thebibliography}
\bibliographystyle{acl_natbib}

\clearpage
\newpage

\appendix

\section{Appendix}
\label{sec:appendix}

\subsection{Experimental Setting Details} \label{ssec::setting-details}
\paragraph{Implementation} For most of our experiments, the language model used in our architecture is the official pre-trained BERT (\textit{bert-base-uncased}) \cite{devlin2019bert} by Hugging Face \cite{wolf-etal-2020-transformers}. Only for the TREC CAR experiment, we used a different pre-trained model as in \citet{khattab2020colbert}, which is the pre-trained \textit{Large} model released by \citet{nogueira2019passage} who ensures that the test set of TREC CAR is not exposed for pre-training.
Since TREC CAR is based on Wikipedia, \citet{nogueira2019passage} pre-trained BERT on the subset of Wikipedia so that the model is not exposed to the test set of TREC CAR during the pre-training step. They released their pre-trained \textit{Large} model and we fine-tune it for TREC CAR experiment. Fine-tuning this model is remarkably slower than BERT \textit{Base}, we leverage the single layer (23) of BERT \textit{Large} to report the result of TREC CAR. 
For the buckets on top of the LM, we modified Winner\text{-}Take\text{-}All (WTA) layer \cite{makhzani2015winner} by adding a weight sparsity proposed by \citet{DBLP:journals/corr/abs-1903-11257}. Our UHD model has about 172M parameters per bucket with $n = 81,920$ for MS MARCO and 398M for TREC CAR. For the binarized inverted index, we use a bit packing compression for memory efficiency. Since compression is not necessary in an environment with sufficient memory, we excluded the time consuming due to compression in the latency measurement.

\paragraph{Training} We use 25\% of MS MARCO train triples, (\textit{query}, \textit{positive passage}, \textit{negative passage}), containing a total of 39,782,779 triples, in all our experiments unless otherwise specified. Note that we use only queries and positive passages with in-batch negatives for training, ignoring the negative passages. We sample 7,000 queries that are not in the above 25\%  to construct a re-ranking train subset with the corresponding top 1000 passages, provided by MS MARCO. This train subset is used for the bucket weight tuning in Section \ref{sec::4.6_multiBucket}. For TREC CAR, we use the first four of five pre-defined folds as a training set which consists of query-passage pairs (approximately 5M pairs), following related works \cite{nogueira2019document,nogueira2019passage,khattab2020colbert}. We utilized dataset pre-processed and provided by \citet{nogueira2019passage}. Training a single-bucket model from the last layer of BERT with $n$ = 81,920 (dimensionality) and $k$ = 80 (non-zero dimensions) takes 15.5 hours on two GeForce RTX 3090 devices for MS MARCO and 40.86 hours on a same single device for TREC CAR.

\paragraph{Evaluation} MS MARCO provides two settings: re-ranking and full ranking. In the re-ranking setting, given a query and 1000 passages (top-1000) retrieved by BM25, the model should re-rank the passages. In full ranking, on the other hand, the model should retrieve relevant passages from the entire collection for a given query. While our main goal is to validate our model in the full ranking setting, we utilize the re-ranking setting when a large number of model variants are compared, as a way to save computational cost. In both settings we use the MS MARCO dev set, which contains 6,980 queries. MRR@10 (mean reciprocal rank) is used as the evaluation metric as in the official evaluation. We use recall at various levels as additional metric. For TREC CAR, we use the same test set (2,254 queries) used to submit to TREC-CAR 2017. Following \citet{nogueira2019document,khattab2020colbert}, we report re-ranking results after BM25 retrieval for fair comparison, using MRR@10 as well as the official metric, MAP (Mean Average Precision). Single RTX 3090 was used for encoding queries and documents, and re-ranking. For full ranking using a binarized inverted index, we use a Threadripper 3960X (CPU) with 24 cores. 

\paragraph{Hyperparameters}
We use the Adam optimizer with the learning rate of 5e-6, $\beta_1$ = 0.9, $\beta_2$ = 0.999 and epsilon = 1e-8, and L2 weight decay of 0.01. The learning rate is linearly warmed up over the first 2,000 steps, and then linearly decayed. The batch size is 32, and the epochs are 0.25 for MS MARCO (25\% of MS MARCO Train Triples Small set) and 1.0 for TREC CAR (all query-passage pairs in the four folds of TREC CAR). For the BERT LM, we use GELU \cite{hendrycks2016bridging} as an activation and set the maximum length to 32 for queries and 180 for passages (truncated if exceeded). For the WTA layer, we select dimensionality $n$ from \{8,192, 24,576, 49,152, 81,920\} and the number of non-zero dimensions $k$ from \{4, 8, 16, 20, 40, 80, 200, 400\}. For the analysis of $n$ and $k$, we once trained with the k of 400 and changed it to above settings at test time. There was no significant difference between the setting where k matches and the setting that does not match at training and test times. After having 3 times of trials for each setting, our final setting is $n = 81,920, k = 80$, matching the balance between the performance (MRR@10) and efficiency. We set weight sparsity to 0.7 from \{0.3, 0.5, 0.7\}, meaning that 70\% of the weights connecting each WTA node to input is set to zero.

\subsection{Analysis on UHD Embeddings}
\label{sec::a2_analysis_embedding}

\begin{figure}
  \centering
  \includegraphics[width=0.9\linewidth]{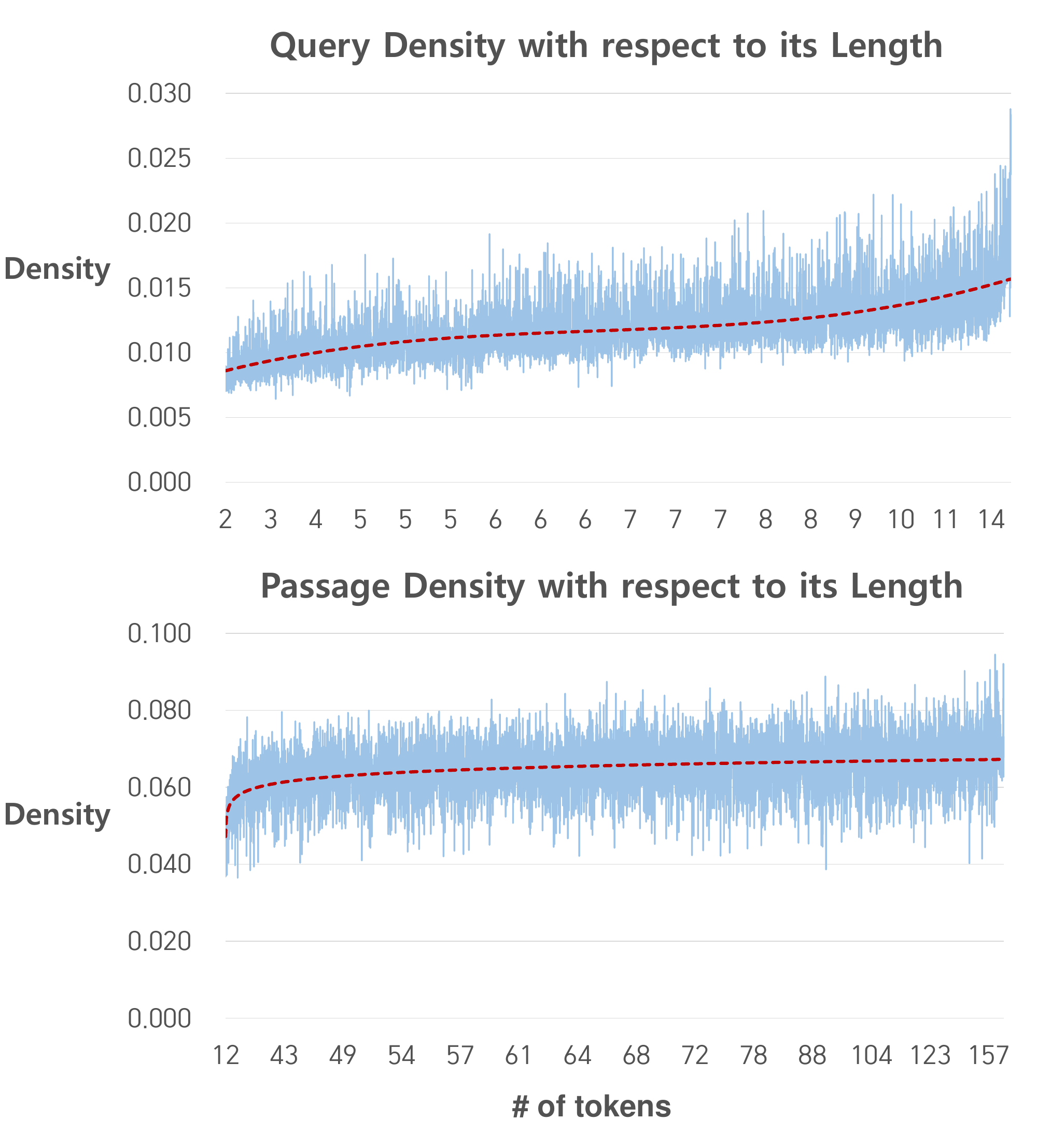}
  \caption{Density of the generated sparse representation of queries and passages. Queries and passages are sorted according to their token lengths (x-axis).}
  \label{fig:sparsity}
\end{figure}

To better understand the characteristics of our proposed model, we conducted an in-depth analysis of the generated embeddings. In this experiment, we use a model with a single bucket on the 12th layer of BERT LM, where dimensionality $n$ and the number of non-zero dimensions $k$ are 81,920 and 80, respectively. 

In order to check whether the generated embedding properly contains the information of queries and passages, in Figure \ref{fig:sparsity}, we measured the density (ratio of non-zero dimension) of the queries (left-hand side) and passages (right-hand side) with respect to their length. Intuitively, we can assume that the more words (or tokens) the text contains, the more information it contains. Because each dimension of sparse embedding is activated only when the corresponding feature exists, the density of the generated embeddings must correlate with the amount of information (the number of tokens) in the query/passage, as shown in Figure \ref{fig:sparsity}. However, interestingly, the density continues to increase according to the number of tokens in the query, while it converges in the passage. Our interpretation is that since the query reflects the intention of the user, it is highly likely that each query term represents a unique feature that narrows down the search space, which does not overlap with other query terms. For the example \textit{'how much do psychologists earn uk'}, a user has put \textit{'uk'} to reduce the search space. As such, increasing the length of the query has a high probability of adding new features, which increases the density. Unlike queries, passages usually deal with limited topics, which makes density converges even if the length increases. 

\begin{figure}[h]
  \centering
  \includegraphics[width=0.9\linewidth]{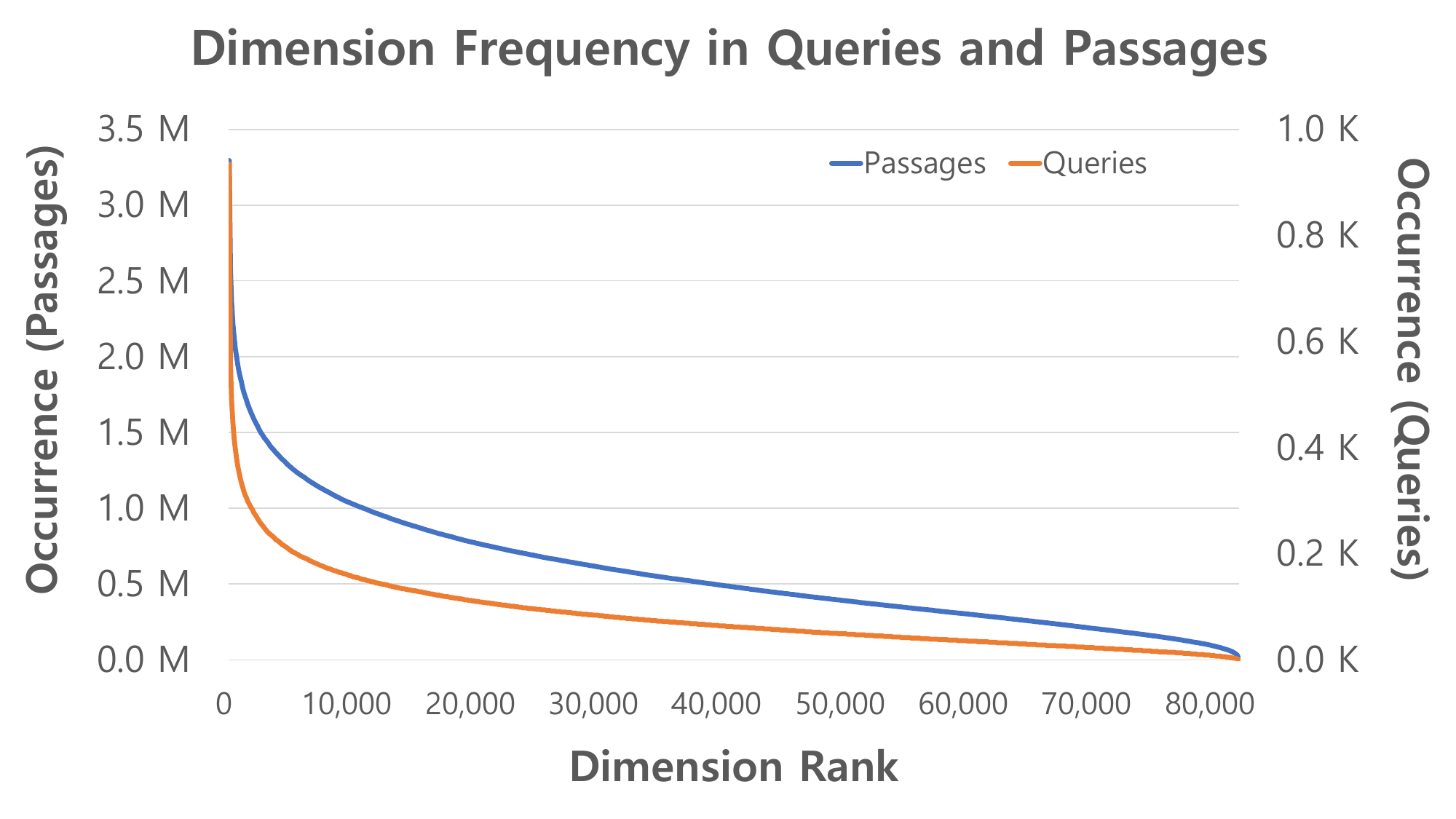}
  \caption{Active (non-zero) dimension frequency in queries and passages.}
  \label{fig:frequency}
\end{figure}

Besides the above query/passage-level embedding analysis, dimension-level analysis is also crucial to understand the characteristics of our proposed model. In Figure \ref{fig:frequency}, we measured how many times each dimension was activated (non-zero) in the query and passage, similar to traditional word frequency analysis. The analysis suggests that the embeddings generated by our model work in a similar fashion to words in natural languages (Zipf's law). It is also worth noting that similar to stop words in natural language words, there are dimensions that appear in many passages (8.84M in total) and queries (7K in total). Dealing with these "stop-dimensions" can also be expected to have a big impact on performance, but we leave it as future work.

\subsection{Additional Analysis on Multiple Buckets}
\label{sec::a4_buckets}

\begin{table}
\begin{center}
\begin{tabular}{lc}
\toprule
\textbf{Strategy} & \textbf{MRR@10}   \\ \midrule 
Single-Bucket \\ ($b=1$, $n=49,152$, $k=480$) & 26.18 \\ 
\hline
Multi-Bucket \\ ($b=6$, $n=8,192$, $k=80$) &  \\ 
\hspace{0.15cm}  Multiple bucket (Joint) & 25.58 \\
\hspace{0.15cm}  Multiple bucket (Separate) & 27.08 \\
\hspace{0.15cm}  + bucket weight tuning & \textbf{27.41} \\
\hline
\hspace{0.15cm}  Ideal query-layer predictor & \textbf{39.62} \\
\bottomrule
\end{tabular}
\end{center}
\caption{Comparison of various multi-bucket strategies on the MS MARCO Passage Retrieval dev set re-ranking task. The total dimensionality ($n$) and non-zero dimension size ($k$) are the same across the strategies. $b$ stands for the number of buckets.}
\label{table:additional-multilayer}
\end{table}

In addition to the method and analysis on multiple buckets (Section \ref{sec::3_4_multibucket}, \ref{sec::4.6_multiBucket}), here we address further various settings and their analysis.

For the multiple bucket setting, we evaluate two types of implementation (Joint/Separate). The joint setting means that we train all WTA layers in an end-to-end manner, while the separate setting denotes that we separately train the model for each WTA layer and combine the results later.
We notice that the joint setting of the multi-bucket unexpectedly underperforms the single-bucket. We speculate that the WTA layers in the joint setting easily interfere with each other during training, implying the need for a special mechanism to handle the interference for future work. On the other hand, the separate setting, which is free from the interference, outperforms the single bucket setting, despite the disadvantages of the multi-bucket strategy (small $n$ for each bucket). This is also evident in the result of the bucket weight tuning. By controlling the weight of each bucket, we can further increase the performance of the multiple bucket setting. 

For the bucket weight tuning, we conducted a grid search on the top-1000 re-ranking dataset we built from the MS MARCO dataset (See Section \ref{ssec::setting-details}), which yields the searched weights ($0.66:0.0:0.33:1.00:0.33:1.00$).

Going further from this analysis, one can imagine that each query may have a more suitable set of layer weights. For example, certain queries may require syntactic contents while others may require topical (semantic) contents. From this perspective, we measured the performance when a layer with the highest MRR was ideally selected for each query (Ideal query-layer predictor in Table \ref{table:additional-multilayer}). The result clearly suggests that selecting different layers or bucket weight distributions for each query has the potential to improve performance. This implies the necessity of a bucket weight predictor which takes a query as an input, but we leave it as future work.

\subsection{Qualitative Analysis on Interpretability}
\label{sec::a_4_interpret}

\begin{table*}[ht!]
\begin{center}
\begin{tabular}{c|l|c}
\toprule
\multirow{2}{*}{\textbf{Dim}} & \multirow{2}{*}{\textbf{Appeared Query Terms}} & \textbf{Inter-} \\
 & & \textbf{pretation} \\
\midrule 
68,604 & \textit{costs}, \textit{paid}, \textit{fee}, \textit{prices}, \textit{fees} & Price \\
29,938 & \textit{tall}, paid, \textit{square}, \textit{feet}, \textit{miles}, \textit{words} & Unit \\
8,797 & \textit{soon}, \textit{longest}, heal, \textit{wait}, \textit{been}, \textit{length}, \textit{weeks}, & Period \\
66,589 & \textit{growth}, \textit{movement}, \textit{development}, degree, theory & Advancement \\
3,740 & \textit{treatment}, \textit{heal}, \textit{surgery}, \textit{doctors}, \textit{doctor} & Treatment \\
24,673 & \textit{medication}, normal, \textit{constipation}, \textit{doctor}, \textit{medicine} & Medical \\
37,089 & \textit{problems}, \textit{diseases}, \textit{signs}, \textit{syndrome}, \textit{fever} & Disease \\
476 & \textit{die}, \textit{killed} & Death \\
47,873 & \textit{technology}, \textit{science}, \textit{chemical}, \textit{scientific}, union & Science \\
374 & science, technology, fastest, \textit{java}, \textit{language}, \textit{software} & Programming \\
68,111 & lower, low, clear, \textit{deductible}, \textit{irs}, depression & Tax \\
21,822 & \textit{party}, beach, \textit{amendment}, tissue, \textit{constitution} & Politic \& Law \\
22,627 & \textit{founded}, \textit{invented}, \textit{begin}, \textit{established}, \textit{started}, released & Founding \\
3,266 & \textit{numbers}, \textit{customer}, \textit{telephone}, \textit{zip}, \textit{contact}, union, npi & E-commerce \\
46,843 & \textit{washington}, \textit{il}, \textit{ma}, \textit{mn}, \textit{indiana}, \textit{south} & Location \\
61,183 & \textit{child}, \textit{kids}, \textit{children}, \textit{disney}, paint, \textit{parents} & Child \\
48,698 & \textit{education}, \textit{student}, training, \textit{degree}, workout & Education \\
26,433 & \textit{synonym}, \textit{description}, \textit{synonyms}, \textit{defined}, english & Definition \\
1,377 & \textit{sang}, \textit{played}, \textit{sings}, requirements, \textit{cats}, \textit{plays} & Play (theatre)\\
\bottomrule
\end{tabular}
\end{center}
\caption{Interpretation of the each dimension in the UHD embedding by analyzing terms of queries that activates a specific dimension. The query terms that support the interpretation are indicated in \textit{italic}.}
\label{table::interpretability}
\end{table*}

One of the advantages of the sparse representation is its interpretability, which is possible in our case due to the disentangled feature of the UHD embedding. To prove this, we conduct an analysis about the meaning of each dimension of the generated UHD embedding. In this analysis, we use the UHD-BERT model with reduced $k(=40)$, to reduce the active dimension for making the analysis easier. 

Table \ref{table::interpretability} shows interpretation results on a few sampled dimensions of the UHD embedding. We use a simple method that validates the interpretation, which is to inspect terms of queries that have non-zero values (activated) on a specific dimension. For example of the dimension index "61,183" in Table \ref{table::interpretability}, the terms that activated queries frequently have were \textit{child}, \textit{kids} and \textit{disney}, implying that the dimension is specialized for the "child" information. For clarity, query terms that have appeared at least five times are listed, along with dimensions that are relatively easy to interpret. We observe that there are dimensions representing abstract latent terms such as "Unit" or "Period". To summarize the analysis, our UHD embedding is not only interpretable (advantage of sparse representations), but also has the power of the latent representation (advantage of dense representations). 

\end{document}